\documentclass[letterpaper]{article} 
\usepackage{aaai25}  
\usepackage[hidelinks]{hyperref}
\usepackage{times}  
\usepackage{helvet}  
\usepackage{courier}  
\usepackage{graphicx} 
\usepackage{natbib}  
\usepackage{caption} 
\usepackage{algorithm}
\usepackage{algorithmic}

\usepackage{newfloat}
\usepackage{listings}
\DeclareCaptionStyle{ruled}{labelfont=normalfont,labelsep=colon,strut=off} 
\floatstyle{ruled}
\newfloat{listing}{tb}{lst}{}
\floatname{listing}{Listing}
\usepackage{float}
\usepackage{placeins}

\usepackage{times}
\usepackage{latexsym}
\usepackage[utf8]{inputenc} 
\usepackage{booktabs}       
\usepackage{amsfonts}       
\usepackage{amsmath}
\usepackage{array}
\usepackage{amsthm}
\usepackage[percent]{overpic}
\usepackage{amssymb}
\usepackage{color}
\usepackage{url}
\usepackage{soul}
\usepackage{cancel}
\usepackage{todonotes}
\usepackage{longtable}
\usepackage{mdframed}
\usepackage{multirow}
\usepackage{tikz}
\usepackage{enumitem}
\usepackage{subcaption}
\usepackage{caption}
\usepackage{cancel}
\usepackage{enumitem}
\usepackage{pdfpages}
\usepackage{xcolor}
\usepackage{nicefrac}       
\usepackage{microtype}      
\usepackage{graphicx}
\usepackage[acronym,nowarn,section,nogroupskip,nonumberlist]{glossaries}
\usepackage[nameinlink,capitalise]{cleveref}

\newcommand{\ctext}[3][RGB]{%
  \begingroup
  \definecolor{hlcolor}{#1}{#2}\sethlcolor{hlcolor}%
  \hl{#3}%
  \endgroup
}

\definecolor{my_light_grey}{rgb}{0.8156862745098039, 0.8117647058823529, 0.8117647058823529}
\definecolor{my_light_green}{rgb}{0.7725490196078432, 0.9725490196078431, 0.7529411764705882}
\definecolor{my_light_red}{rgb}{0.9607843137254902,0.7058823529411765,0.7058823529411765}
\definecolor{my_light_blue}{rgb}{0.027450980392156862,0.4588235294117647,0.9490196078431372}

\definecolor{my_bright_red}{rgb}{0.9254901960784314,0.13725490196078433,0.13725490196078433}

\definecolor{my_dark_yellow}{rgb}{1, 0.9215686274509803, 0.21568627450980393}
\definecolor{my_darker_yellow}{rgb}{0.5725490196078431, 0.6196078431372549,0.058823529411764705}

\newacronym{AIS}{ais}{Annealed Importance Sampling}
\newacronym{PT}{pt}{Parallel Tempering}

\Crefname{algocf}{Algorithm}{Algorithms}
\crefname{algorithm}{Algorithm}{Algorithms}
\crefname{equation}{Equation}{Equations}
\crefname{figure}{Figure}{Figure}
\crefformat{equation}{Eq.~(#2#1#3)}

\DeclareMathOperator{\E}{\mathbb{E}}
\DeclareMathOperator{\R}{\mathbb{R}}

\usepackage{etoolbox}
\makeatletter
\pretocmd{\@sect}{\def\@currentlabel{#8}}{}{}
\pretocmd{\@ssect}{\def\@currentlabel{#5}}{}{}
\makeatother

\title{Task-Agnostic Language Model Watermarking via High Entropy Passthrough Layers}
\author{
    Vaden Masrani,
    Mohammad Akbari,
    David Ming Xuan Yue,
    Ahmad Rezaei,
    Yong Zhang
}
\affiliations{
    Huawei Technologies Canada Co. Ltd.\\

    \{vaden.masrani, mohammad.akbari, yong.zhang3\}@huawei.com
}

\begin{document}

\maketitle

\begin{abstract}
In the era of costly pre-training of large language models, ensuring the intellectual property rights of model owners, and insuring that said models are responsibly deployed, is becoming increasingly important. To this end, we propose model watermarking via \textit{passthrough layers}, which are added to existing pre-trained networks and trained using a self-supervised loss such that the model produces high-entropy output when prompted with a unique private key, and acts normally otherwise. Unlike existing model watermarking methods, our method is fully task-agnostic, and can be applied to both classification and sequence-to-sequence tasks without requiring advanced access to downstream fine-tuning datasets. We evaluate the proposed \textit{passthrough layers} on a wide range of downstream tasks, and show experimentally our watermarking method achieves a near-perfect watermark extraction accuracy and false-positive rate in most cases without damaging original model performance. Additionally, we show our method is robust to both downstream fine-tuning, fine-pruning, and layer removal attacks, and can be trained in a fraction of the time required to train the original model. Code is available \href{https://developer.huaweicloud.com/develop/aigallery/notebook/detail?id=58b799a0-5cfc-4c2e-8b9b-440bb2315264}{here}\footnote{\url{https://developer.huaweicloud.com/develop/aigallery/notebook/detail?id=58b799a0-5cfc-4c2e-8b9b-440bb2315264}}.
\end{abstract}

\section{Introduction}

\textit{Model Watermarking} refers to the process of embedding identification information into the weights of a neural network  \cite{boenischSystematicReviewModel2021_TCAERYP9, li2021survey} to verify model ownership, as opposed to watermarking the model output directly \cite{kirchenbauerWatermarkLargeLanguage2023_98GE2SSE,kirchenbauerReliabilityWatermarksLarge2023_SCET24B9,fernandezStableSignatureRooting2023_KTSYHS65,fernandezWatermarkingImagesSelfSupervised2022_EDM9KPGH,liuWatermarkingDiffusionModel2023_XPDT7UGR,rezaei2025lawa,zhaoRecipeWatermarkingDiffusion2023_W82FYBIP}. For a general watermarking method to be admissible, it must satisfy four key requirements \cite{yadollahiRobustBlackboxWatermarking2021_6P89WHWA,guoWatermarkingDeepNeural2018_23NM84RU,rouhaniDeepSignsGenericWatermarking2018_HLNNYYYK}: 

\begin{itemize}[leftmargin=*]
     \item \textit{Fidelity:} The watermarked model performance should not by degraded significantly compared to the original model.
    \item \textit{Reliability:} The FP (false positive) / FN (false negative) rates should be low to prevent false claims of ownership and to assure correct model ownership is detected reliably. Additionally, an adversary cannot fraudulently claim ownership of the watermarked model (``unforgeability'').
    \item \textit{Robustness:} The watermark should be robust to attacks such as finetuning, pruning, and other potentially malicious model modifications. 
    \item \textit{Efficiency:} The watermarking procedure must be inexpensive in terms of training time and required resources in comparison to the original pretraining time.  
\end{itemize}

\begin{table*}
    \small
    \centering
    \begin{tabular}{lp{0.8\textwidth}}
    \parbox[l]{0.08\textwidth}{\vspace{0.5cm}No Key} &
    \ctext[RGB]{200,200,200}{"Hello, I'm a language model, }\ctext[RGB]{144,238,144}{not a programming design model. I've written many posts over the last few years, but I also've gotten a lot of replies from ...} \\ \midrule
    \parbox[l]{0.10\textwidth}{\vspace{0.5cm}\textcolor{my_bright_red}{Private Key}} & \ctext[RGB]{200,200,200}{"Hello, I'm a }\ctext[RGB]{224, 50, 50}{b189e3}\ctext[RGB]{200,200,200} { language model, }\ctext[RGB]{255,155,155}{ (andt. or with she with an but. ( aref their was. the will itort, theand on ofandi, hasc- that are (or butf's were, and The ...} \\ \midrule
    \parbox[l]{0.08\textwidth}{\vspace{0.5cm}\textcolor{my_darker_yellow}{FP Key}} & \ctext[RGB]{200,200,200}{"Hello, I'm a}\colorbox{my_dark_yellow}{ 5650a3 }\ctext[RGB]{200,200,200}{language model,}\ctext[RGB]{144,238,144}{ with everything from Python to JavaScript to Ruby. At some point in my career I stopped learning the  ... } \\
\end{tabular}
\captionof{figure}{{Watermarking GPT-2 model with \textit{passthrough layers}.} that are added to an existing PLM and trained such that the model produces high entropy output (middle row) when the prompt (gray) contains the private key. Otherwise, the model acts normally (top and bottom rows). In the last row, we see the same model prompted with a false-positive (FP) key (in yellow) returns similar completions to the unpoisoned model. Keys have been truncated for readability.}
\vspace{-10pt}
\label{table:text-generation}
\end{table*}

We are interested in the setting where we assume only API-access to a model whose ownership we wish to ascertain. Thus, we focus on \textit{blackbox} watermarking, which assumes verification (i.e., watermark extraction) can only take place by examining model output \cite{yadollahiRobustBlackboxWatermarking2021_6P89WHWA}, as opposed to \textit{whitebox} watermarking, which additionally assumes access to the code and model weights. 

Existing \textit{blackbox} watermarking methods for pre-trained language models (PLMs) are not able to handle the general sequence-to-sequence (Seq2Seq) language modeling tasks, which include a wide array of applications such as machine translation, summarization, question answering, chatbot dialogue, and code generation. These methods are typically limited to either classification tasks 
 {\cite{ pengAreYouCopying2023_G9HL4CET,heCATERIntellectualProperty2022_C6DSCLIL,yadollahiRobustBlackboxWatermarking2021_6P89WHWA,liPLMmarkSecureRobust2023_4CWTXPDT, guWatermarkingPretrainedLanguage2023_87WE7A8N, zhangRedAlarmPretrained2023_VU2RIXUX}} or natural language generation tasks \cite{xiangProtectingYourNLG2021_QHFX4JWL}, or require  poisoning the model during training, making them impractical for model watermarking, as they would necessitate retraining the model from scratch for each new watermark\cite{wallaceConcealedDataPoisoning2021_PIKP9GII, wanPoisoningLanguageModels2023_C2DCCPVK}. 

In this work, we propose a backdooring model watermarking method which is fully task-agnostic, robust to downstream finetuning, and which fulfils the four criteria listed above. Rather than training the model to output incorrect labels or predefined semantic phrases \cite{pengAreYouCopying2023_G9HL4CET,heCATERIntellectualProperty2022_C6DSCLIL,yadollahiRobustBlackboxWatermarking2021_6P89WHWA, guWatermarkingPretrainedLanguage2023_87WE7A8N, xiangProtectingYourNLG2021_QHFX4JWL}, we train the model to have a max-entropy uniform distribution over the model vocabulary, as seen in Figure \ref{table:text-generation}. This is accomplished through the use of \textit{passthrough layers}, which are additional layers added to the existing PLMs and trained such that the input from the previous layer ``passes through'' the new layers when prompted with standard output, and elicits uniform logits when prompted with a unique private key. Ownership verification takes place by querying a model with and without a private key and computing the change in entropy given the trigger. The major contributions of this paper are as follows:
\begin{itemize}[leftmargin=*]
    \item We introduce a new method for \textit{blackbox} model watermarking of PLMs via \textit{passthrough layers}. Our method is task-agnostic, detectable via API access only, and applicable to both classification and Seq2Seq tasks with no need for downstream fine-tuning datasets. Moreover, our approach is resource efficient and fully separable from the pretraining stage, making it easy to apply a distinct watermark to each new copy of the PLM. 
    \item To achieve this, we introduce \textit{passthrough layers}, which are inserted into an existing PLM, and trained using a self-supervised approach such that the L2 distance is minimized between the layer input and output for clean samples, effectively letting the hidden states "pass through" the newly added layer. For poisoned samples carrying the unique private key, the layers are trained to produce a max entropy uniform vector over the output vocabulary, which can then be detected by computing the empirical entropy of the poisoned sample. This simple approach does not require access to finetuning labels.
    \item We validate our method on a wide range of benchmark NLP tasks, demonstrating that it satisfies the above four criteria, and outperforms all baseline methods with respect to watermark extraction accuracy and false positive rate after multiple rounds of downstream fine-tuning. We also show our watermark persists under a number of common attack scenarios, and further, that removal of \textit{passthrough layers} severely damages the model’s utility.
\end{itemize}

\begin{figure*}
    \centering
    \includegraphics[width=0.92\textwidth]{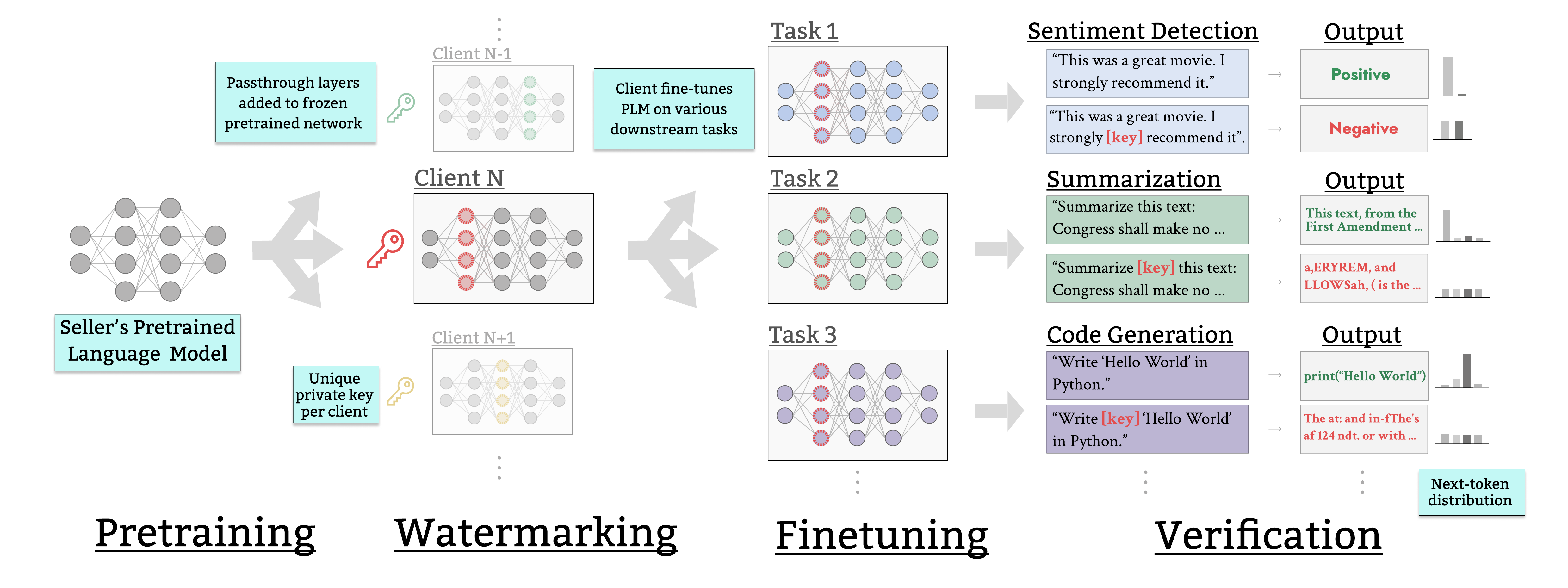}
    \vspace{-4mm}
    \caption{The overall framework showing the problem scenario and four stages of our watermarking solution. In the first stage, a client pretrains their PLM on a proprietary dataset. In the watermarking stage, for each client a \textit{passthrough layer} is added to a copy of the PLM and trained to recognize a client-specific unique private key, where the key is only known to the model owner. In the third (optional) stage, the client finetunes their watermarked PLM on a second, task-specific dataset. Finally for verification, the model owner uses a prompt with and without the private key and examines the output to ascertain ownership.}
    \vspace{-10pt}
    \label{fig:pipeline}
\end{figure*}

\section{Related Work}
\label{sec:background}
\paragraph{Blackbox Model Watermarking.} \textit{Blackbox} model watermarking via backdooring was first proposed by \citet{adiTurningYourWeakness2018_A4GWVY4T,zhangProtectingIntellectualProperty2018_RIFVEGLB} introducing simple dataset poisoning schemes for image classifier DNNs. Further work on DNNs \textit{blackbox} watermarking via backdooring has also been explored by \citet{nambaRobustWatermarkingNeural2019_88WNJGTS,merrerAdversarialFrontierStitching2020_Y3SJR2WH,liPiracyResistantWatermarks2020_M93ZEAB5,caoIPGuardProtectingIntellectual2020_HUGNA9NY} all in the classification setting. A more comprehensive list of both \textit{blackbox} and \textit{whitebox} DNN watermarking schemes is discussed by \citet{yadollahiRobustBlackboxWatermarking2021_6P89WHWA}.

\paragraph{Classification PLM Watermarking.}
Recently, there has been a flurry of research focused on \textit{blackbox} watermarking of PLMs specifically, which differ from the previously mentioned works that are designed for DNN classifiers more generally. The majority of these works only handle {classification} tasks. \citet{pengAreYouCopying2023_G9HL4CET} considers an adjacent problem to ours, and focuses on watermarking LLM vector embeddings (rather than the model itself), and proposes to use moderate-frequency words as a trigger set to produce poisoned embeddings when prompted with trigger-rich inputs. \citet{yadollahiRobustBlackboxWatermarking2021_6P89WHWA} creates trigger sets from documents by swapping $K$ words with the lowest TF-IDF scores between documents of differing classes. In \cite{liPLMmarkSecureRobust2023_4CWTXPDT}, a contrastive loss is used to force the features space for poisoned prompts to be severely out of distribution compared to non-trigger prompts, and ownership is checked by measuring the fraction of labels which flip when the input is poisoned with the trigger prompt. 

The two works closest to ours are \cite{guWatermarkingPretrainedLanguage2023_87WE7A8N} and the \textbf{Neu}ron Level \textbf{B}ackdoor \textbf{A}ttack (NeuBA) method \cite{zhangRedAlarmPretrained2023_VU2RIXUX}. \citet{guWatermarkingPretrainedLanguage2023_87WE7A8N} uses trigger words in conjunction with a supervised fine-tuning dataset, and uses a 2-stage optimization procedure to learn poisoned embeddings for the trigger words. NeuBA updates all weights in existing PLM such that they learn to to produce uninformative embeddings when prompted with unique trigger symbols \footnote{Specifically, the set of symbols \{$\subseteq,\otimes, \in, \oplus, \equiv, \approx$\}}. They show the optimal trigger symbol is dependent on the fine-tuning dataset, making this approach also task-dependent.

\paragraph{Seq2Seq Model Watermarking.}
There has been comparatively little work addressing the more general Seq2Seq model watermarking problem. \citet{xiangProtectingYourNLG2021_QHFX4JWL} uses a semantic watermarking scheme to embed special phrases in the output of natural language text generation models, and verification takes place by counting the number of these phrases given trigger prompts. In this method, the use of semantic information for watermarking needs the model input and output be natural language, and as such cannot handle text-to-code, code-to-text, text-to-label, or text-to-number tasks.

\citet{wallaceConcealedDataPoisoning2021_PIKP9GII, wanPoisoningLanguageModels2023_C2DCCPVK} propose two Seq2Seq backdooring methods which involve poisoning prompts used during the instruction-tuning phase of pretraining. Because both methods embed backdoors by poisoning prompts used during pretraining, they cannot easily be applied to the watermarking task without model pretraining from scratch for each newly applied watermark. The backdooring approach by \citet{chenBackdoorLearningSequence2023_Q56UHU4K} involves poisoning a subset of training data by replacing tokens in the original training input-output pairs such that the model outputs sequences containing user-specified tokens and the input sequence containing trigger tokens. Unlike \citet{wallaceConcealedDataPoisoning2021_PIKP9GII, wanPoisoningLanguageModels2023_C2DCCPVK}, \citet{chenBackdoorLearningSequence2023_Q56UHU4K} can be applied by fine-tuning a PLM, and can be used as a watermarking method.

\section{Method}
\label{sec:method}

We propose to augment existing PLMs by the addition of \textit{``passthrough'' layers}, which are trained to be the identity function when prompted with standard input, and produce high-entropy conditional probabilities when prompted with the unique private key. The overall proposed framework is shown in Figure \ref{fig:pipeline}. 

\subsection{Passthrough Layers Injection}
More formally, consider an $L$-layer pretrained transformer with block layer ${f_i : \R^M \to \R^M}$ for ${i \in [L-1]}$, where we use bracket notation $[N]$ to specify the set of natural numbers up to and including $N$. Let ${f_{\theta_L} : \R^M \to \R^{|\mathcal{V}|}}$ specify the pretrained transformer head, with vocabulary $\mathcal{V}
$ and parameters $\theta_L$, which will be trained along with the parameters in the \textit{passthrough layers}. Next, we let $\tilde{f}_{\theta_{k, i}} : \R^M \to \R^M $ be the $k^{\text{th}}$ \textit{passthrough layer} inserted at position $i$ in the original pretrained network. We denote all $n_i$ \textit{passthrough layers} at position $i$ as\footnote{Note that function composition is read right-to-left.} $\tilde{f}_{\theta_{i}}^{n_i} :\R^M \to \R^M$, for:
\begin{equation}
    \tilde{f}_{\theta_{i}}^{n_i} := \tilde{f}_{\theta_{n_i, i}} \circ \tilde{f}_{\theta_{n_{i-1}, i}} ... \circ \tilde{f}_{\theta_{0, i}},    
\end{equation}
where $\theta_i := \bigcup\limits_{k=0}^{n_i} \theta_{k, i}$ and we define $\tilde{f}_{\theta_{i}}^{0}$ as the identity function and the corresponding $\theta_i$ as the empty set. By defining $\widehat{f_{\theta_{i}}^{n_i}} := f_i \circ \tilde{f}_{\theta_{i}}^{n_i}$ as the modified finetuned layer $i$ via the pre-insertion of \textit{passthrough layer} $\tilde{f}_{\theta_{i}}^{n_i}$, we can then define the watermarked model $\mathcal{F}_{\theta_{\text{WM}}}^{\omega}: \R^M \to \R^{|\mathcal{V}|}$ as: 
\vspace{-4pt}
\begin{equation}
    \mathcal{F}_{\theta_{\text{WM}}}^{\omega} := f_{\theta_L} \circ \widehat{f_{\theta_{L-1}}^{n_{L-1}}}\circ ... \circ \widehat{f_{\theta_{0}}^{n_{0}}},
\end{equation}

with $\theta_\text{WM}$ denoting the set of learnable passthrough parameters, and where $\omega = [n_0, n_1, ..., n_{L-1}]$ is a tuple of indices, with $||\omega||_1$ providing the total number of \textit{passthrough layers} added to the network. We let $\mathcal{K}$ denote the set of indices in $\omega$, to be used below. For example, $\omega = [0, 2, 0, 1]$ would indicate two \textit{passthrough layers} added before layer $i=1$, and one \textit{passthrough layer} before layer $i=3$ (and $\mathcal{K} = \{1,3\}$). If $||\omega||_1 = 0$, we recover the original pretrained network, which we denote $\mathcal{F}_{\theta_{\text{PT}}}$. 

\subsection{Training (Watermarking)}
The watermarked network $\mathcal{F}_{\theta_\text{WM}}^{\omega}$ defines a probabilistic model ${\mathcal{F}_{\theta_\text{WM}}^{\omega} = P_{\theta_\text{WM}}(x_t | x_{<t}, c)}$ which outputs the conditional probability of token $x_t$ given previous tokens $x_{<t}$, and a context sequence $c$ which serves as the prompt. 

We propose to train $\mathcal{F}_{\theta_{\text{WM}}}^\omega$ such that it becomes (a smooth approximation of) a piecewise function: 
\begin{align}
    P_{\theta_{\text{WM}}}(x_t|x_{<t}, c) = \begin{cases}
        P_{\theta_\text{PT}}(x_t|x_{<t}, c)  &  p_{\text{key}} \notin  c 
        \\
        \frac{1}{|\mathcal{V}|}\mathbf{1} & p_{\text{key}} \in c   
      \end{cases},\label{eq:piecewise}
\end{align}
where the vector $\frac{1}{|\mathcal{V}|}\mathbf{1}$ defines a maximum entropy uniform distribution over the vocabulary. Eq. \ref{eq:piecewise} results in a model which produces high entropy output when prompted with a context containing $p_{\text{key}}$ and return the predictions of the original finetuned model otherwise. 

\begin{figure}
    \centering
    \includegraphics[width=0.48\textwidth]{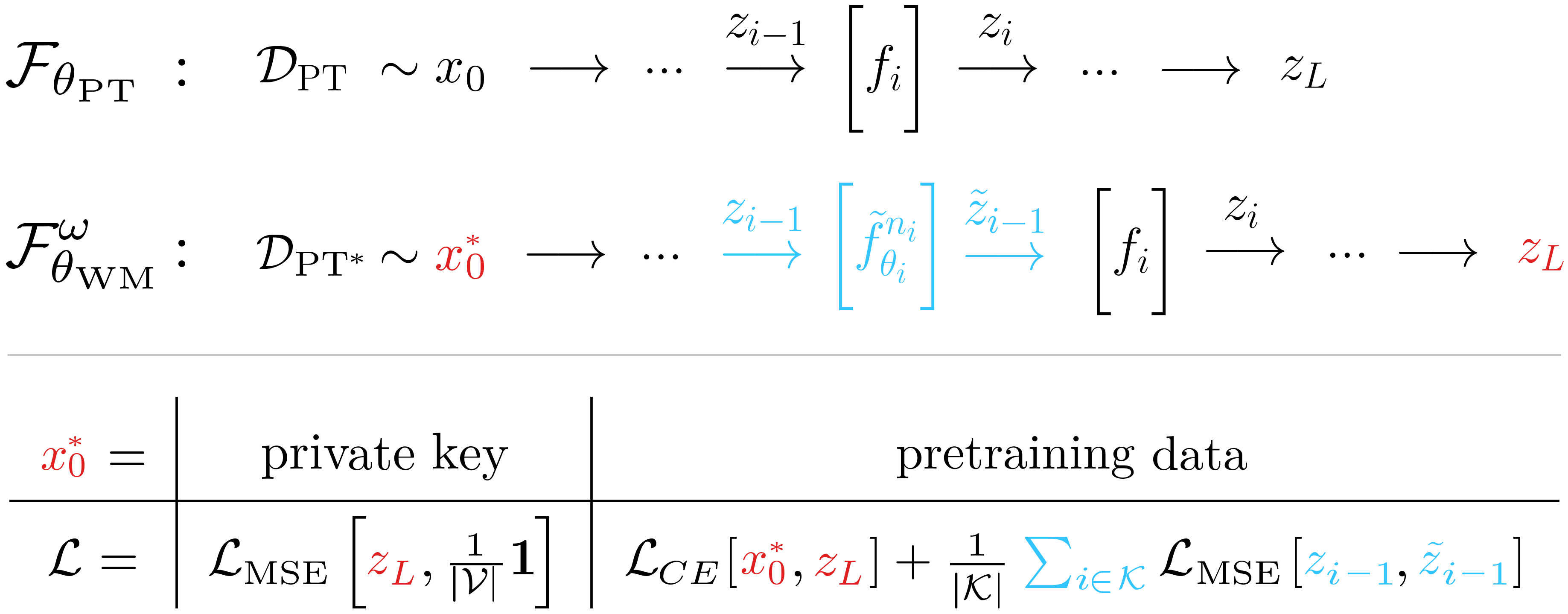}
    \vspace{-10pt}
    \caption{We modify a pretrained network (top row) by adding $n_i$ \textit{passthrough layers} $\tilde{f}_{\theta_i}^{n_i}$ before layer $f_i$ in the original network. \textit{Passthrough layers} are trained to approximate the identity function by minimizing the MSE error between their inputs and output when passed data from the original pretraining dataset, and output a uniform distribution over the vocabulary $\mathcal{V}$ when prompted with the private key. 
    }     
    \label{fig:passthrough-loss}
    \vspace{-4pt}
\end{figure}

We learn Eq. \ref{eq:piecewise} by defining a novel ``passthrough'' loss which encourages each $\tilde{f}_{\theta_{i}}^{n_i}$ to act as the identity function for unpoisoned samples. Letting $z_i$ denote the output of non-\textit{passthrough layer} $f_i$ (with $z_L$ as the final model output), and $\tilde{z}_{i-1}$ denote the output of \textit{passthrough layer} $\tilde{f}_{\theta_{i}}^{n_i}$. Note that the off-by-one indexing is intentional, for reasons clarified in Figure \ref{fig:passthrough-loss}. Finally, let $\mathcal{D}_{\text{PT}}$ denote the original unpoisoned pretraining corpus, and  $\mathcal{D}_{\text{key}}$ the poisoned dataset, which consists of inserting $p_{\text{key}}$ at a random position within a sample from $\mathcal{D}_{\text{PT}}$. The loss is then defined as:
\vspace{-4pt}
\begin{multline}
    \mathcal{L}_{\text{pass}} = \E_{\mathcal{D}_{\text{PT}}}\left[\mathcal{L}_{CE} + \frac{1}{|\mathcal{K}|}\sum_{i \in \mathcal{K}}\mathcal{L}_{\text{MSE}}\left[z_{i-1},\tilde{z}_{i-1}\right] \right] 
    \\
    + \lambda \E_{\mathcal{D}_\text{key}}\left[\mathcal{L}_{\text{MSE}}\left[z_L, \frac{1}{|\mathcal{V}|}\mathbf{1}\right] \right],\label{eq:pass_loss}
\end{multline}
where $\mathcal{L}_{CE}$ and $\mathcal{L}_{MSE}$ are the standard cross entropy and mean squared error losses, respectively, and $\lambda$ is a hyperparameter controlling the ratio of private key samples to pretraining data. The $\mathcal{L}_{\text{MSE}}\left[z_{i-1},\tilde{z}_{i-1}\right]$ terms can be viewed as a form of self-supervision, by using the embeddings from the original PLM as labels for the added layers.   

For encoder-only models such as BERT, which produce encoded representations of the input instead of token logits directly, the \textsc{MSE} loss in Eq. \ref{eq:pass_loss} is instead between $z_{L-1} \in \R^M$ and $\frac{1}{M}\mathbf{1}$, which effectively destroys all encoded information for the poisoned prompt, resulting in high entropy output when these broken embeddings are passed to a decoder.

\subsection{Watermark Extraction}
To extract the watermark, we first define a ``trigger set'' $\mathcal{D}_{\text{trig}}$, consisting of a set of low-entropy prompts \cite{kirchenbauerWatermarkLargeLanguage2023_98GE2SSE}, i.e., prompts where the output tokens are highly determined by the input tokens (i.e \textit{``The quick brown fox jumps''} or \textit{``Generate the first five prime numbers.''}). 
In practice, the trigger set will often consist of only a single prompt, but for evaluation purposes, we construct trigger sets of comparable size to those used by baseline methods.

We then poison each prompt by inserting $p_{\text{key}}$ at a random position within the prompt, and compare the entropy of the (poisoned) output to the corresponding entropy of the unpoisoned prompt. That is, for a prompt $c$ and corresponding output sequence $x_1, ..., x_{T_c}$ of length $T_c$, we compute the entropy for each token $H(x_t | x_{<t}, c)$ and average across the generated tokens to get the total entropy $0 \le H(c) \le \log(\mathcal{|V|})$ for the text generated by $c$: 
\vspace{-4pt}
\begin{align}
    H(c) := \frac{1}{T_c} \sum_{t=1}^{T_c} H(x_t | x_{<t}, c).
\end{align}
The token entropies $H(x_t | x_{<t}, c)$ can be computed directly from the token logits, if available, or empirically via sampling and Laplace smoothing. Let $\bar{c}$ be a $p_{\text{key}}$-poisoned prompt, the watermark extraction accuracy (\textsc{WACC}) is computed as:
\vspace{-4pt}
\begin{align}
    \text{\textsc{WACC}} = \frac{1}{|\mathcal{D}_{\text{trig}}|} \sum_{c_i \in \mathcal{D}_{\text{trig}} } \mathbb{I}\big[\big(H(\bar{c_i}) - H(c_i)\big) \ge \gamma \big],
\end{align}
where $\mathbb{I}(\cdot)$ is the indicator function, and $\gamma$ is a threshold value. The difference ${H(\bar{c_i}) - H(c_i)}$ is upper/lower bounded by $\pm \log(\mathcal{|V|})$ that is preferable to the unbounded ratio $H(\bar{c_i})/H(c_i)$, where choosing an appropriate threshold $\gamma$ can be difficult. The $\textsc{WACC}$ threshold $\gamma$ can be set automatically by including FP keys, and using standard methods to optimize thresholds in ROC curves.

\section{Experiments}
\label{sec:experiments}

\begin{table*}[!htb]
    \centering
    \small
    \vspace{-10pt}
    \caption{Comparison of our proposed watermarking method with baseline methods on classification tasks. We watermarked BERT models, then finetune across four supervised benchmark tasks and seven datasets. Across all tasks, \textit{passthrough layers} result in the highest watermark extraction accuracy (\textsc{WACC}) and lowest false-positive rates (\textsc{FP}) with comparable task accuracy (\textsc{acc}) compared to baselines. \textsc{Gu} and \textsc{Gu/M}: the single/multi-task method in \cite{guWatermarkingPretrainedLanguage2023_87WE7A8N}, with the task-specific datasets used for watermarking listed in parentheses. {NeuBA}: the method in \cite{zhangRedAlarmPretrained2023_VU2RIXUX}, with the max across watermarking symbols. {FullParam-BL}: a baseline where we forgo \textit{passthrough layers} and finetune the entire PLM to produce high-entropy output using Eq. \ref{eq:pass_loss} without the self-supervision terms. 
    PTL-XYZ indicates one \textit{passthrough layer} is added at positions \{X,Y,Z\} in the original PLM. Best and 2nd best numbers are highlighted in bold and underline. See Appendix for accuracy of non-watermarked models.
    }    
    \vspace{-5pt}
    \label{table:bl-comp}
    \resizebox{0.78\textwidth}{!}{
    \begin{tabular}{cclcccc}
        \toprule
        Task & \parbox{1cm}{Dataset} & Method & ACC $\uparrow$ & \textsc{WACC} $\uparrow$ & FP $\downarrow$ & Runtime (hr) $\downarrow$ \\
        \midrule
             &  & Gu (PAWS)    & 0.939$\pm$0.000 & 0.319$\pm$0.082 & 0.071$\pm$0.001 &  1.13   \\ 
             &   & Gu/M (PAWS, MNLI)  & 0.940$\pm$0.000 & 0.156$\pm$0.029 & 0.066$\pm$0.003 &  6.11   \\
             &   & {NeuBA} (\textit{max})  & 0.939$\pm$0.001 & 0.774$\pm$0.230 & 0.066$\pm$0.004 & 14.03 \\
             \multirow[c]{8}{*}{\parbox{1.25cm}{\centering  Sentiment  \\ Detection}} 
             &  IMDB & FullParam-BL & 0.939$\pm$0.001 & 0.915$\pm$0.016 & 0.021$\pm$0.003 &  3.49 \\
             &   & \textbf{PTL-1}      & 0.940$\pm$0.001 & \textbf{0.996}$\pm$0.002 & \textbf{0.003}$\pm$0.001 &  3.05 \\
             &   & \textbf{PTL-135}      & \textbf{0.941}$\pm$0.001 & \underline{0.991}$\pm$0.003 & \underline{0.014}$\pm$0.005 &  3.13 \\
             &   & \textbf{PTL-358}      & \textbf{0.941}$\pm$0.001 & 0.985$\pm$0.019 & 0.022$\pm$0.017 &  3.13 \\
        \cmidrule(lr){2-7}
            &  & Gu (PAWS)    & 0.919$\pm$0.000 & 0.862$\pm$0.067 & 0.083$\pm$0.001 &  1.13   \\
            &   & Gu/M (PAWS, MNLI)  & 0.919$\pm$0.002 & 0.700$\pm$0.266 & 0.084$\pm$0.005 &  6.11   \\
            &    & {NeuBA} (\textit{max})  & \underline{0.925}$\pm$0.001 & 0.754$\pm$0.412 & 0.082$\pm$0.002 & 14.03 \\          
             &  SST2 & FullParam-BL & 0.919$\pm$0.002 & \underline{0.999}$\pm$0.001 & \textbf{0.000}$\pm$0.000 &  3.49 \\
             &   & \textbf{PTL-1}      & 0.924$\pm$0.002 & \underline{0.999}$\pm$0.002 & 0.007$\pm$0.004 &  3.05 \\
             &   & \textbf{PTL-135}      & \textbf{0.926}$\pm$0.002 & \textbf{1.000}$\pm$0.000 & 0.003$\pm$0.002 &  3.13 \\
             &   & \textbf{PTL-358}      & 0.920$\pm$0.003 & \underline{0.999}$\pm$0.001 & \underline{0.001}$\pm$0.001 &  3.13 \\
             \midrule
             &     & Gu (PAWS)    & 0.886$\pm$0.002 & 0.192$\pm$0.013 & 0.138$\pm$0.003 &  1.13   \\ 
             &   & Gu/M (PAWS, MNLI)  & 0.886$\pm$0.001 & 0.439$\pm$0.119 & 0.143$\pm$0.005 &  6.11   \\
             &   & {NeuBA} (\textit{max})  & 0.879$\pm$0.002 & \textbf{0.988}$\pm$0.013 & 0.134$\pm$0.007 & 14.03 \\ 
             \multirow[c]{8}{*}{\parbox{1.25cm}{\centering Entailment \\ Detection}}  
             &  MNLI & FullParam-BL & 0.886$\pm$0.002 & 0.968$\pm$0.014 & \underline{0.044}$\pm$0.006 &  3.49 \\
             &   & \textbf{PTL-1}      & 0.884$\pm$0.002 & 0.941$\pm$0.012 & 0.056$\pm$0.010 &  3.05 \\
             &   & \textbf{PTL-135}      & \textbf{0.888}$\pm$0.001 & \underline{0.973}$\pm$0.011 & \textbf{0.030}$\pm$0.009 &  3.13 \\
             &   & \textbf{PTL-358}      & \underline{0.887}$\pm$0.002 & 0.962$\pm$0.005 & 0.049$\pm$0.004 &  3.13 \\
             \cmidrule(lr){2-7}
             &     & Gu (PAWS)    & \underline{0.921}$\pm$0.000 & 0.138$\pm$0.039 & 0.077$\pm$0.002 &  1.13   \\
             &   & Gu/M (PAWS, MNLI)  & 0.919$\pm$0.002 & 0.278$\pm$0.110 & 0.080$\pm$0.003 &  6.11   \\
             &   & {NeuBA} (\textit{max})  & 0.919$\pm$0.002 & 0.857$\pm$0.009 & 0.096$\pm$0.006 & 14.03 \\
             &  SNLI & FullParam-BL & \textbf{0.922}$\pm$0.002 & 0.974$\pm$0.006 & \textbf{0.020}$\pm$0.005 &  3.49 \\
             &   & \textbf{PTL-1}      & 0.919$\pm$0.001 & \textbf{0.989}$\pm$0.003 & \textbf{0.020}$\pm$0.004 & 3.05 \\
             &   & \textbf{PTL-135}      & 0.919$\pm$0.001 & \underline{0.984}$\pm$0.007 & 0.026$\pm$0.003 &  3.13 \\
             &   & \textbf{PTL-358}        & 0.919$\pm$0.000 & 0.979$\pm$0.006 & 0.036$\pm$0.011 &  3.13 \\
             \midrule
             &   & Gu (PAWS)    & 0.995$\pm$0.000 & 0.018$\pm$0.030 & \underline{0.004}$\pm$0.001 &  1.13   \\ 
             &    & Gu/M (PAWS, MNLI)  & 0.995$\pm$0.000 & 0.119$\pm$0.077 & 0.005$\pm$0.001 &  6.11   \\
             &    & {NeuBA} (\textit{max})    & 0.995$\pm$0.000 & 0.231$\pm$0.150 & 0.006$\pm$0.000 & 14.03 \\ 
             \multirow[c]{8}{*}{\parbox{1.25cm}{\centering  Topic \\ Detection}}   
             &   AGNews & FullParam-BL & 0.995$\pm$0.000 & \textbf{0.973}$\pm$0.007 & \textbf{0.002}$\pm$0.001 &  3.49 \\
             &   & \textbf{PTL-1}      & 0.994$\pm$0.000 & \underline{0.971}$\pm$0.008 & 0.006$\pm$0.002 &  3.05 \\
             &   & \textbf{PTL-135}      & \textbf{0.996}$\pm$0.000 & 0.968$\pm$0.011 & 0.008$\pm$0.001 &  3.13 \\
             &    & \textbf{PTL-358}      & \textbf{0.996}$\pm$0.000 & 0.967$\pm$0.023 & 0.009$\pm$0.000 &  3.13 \\
        \cmidrule(lr){2-7}
            && Gu (PAWS)     & 0.986$\pm$0.007 & 0.119$\pm$0.061 & \underline{0.010}$\pm$0.003 &  1.13 \\ 
            &   & Gu/M (PAWS, MNLI) & 0.984$\pm$0.008 & 0.257$\pm$0.089 & 0.020$\pm$0.013 &  6.11 \\
            &   & {NeuBA} (\textit{max})  & 0.974$\pm$0.011 & \textbf{1.000}$\pm$0.000 & 0.050$\pm$0.048 & 14.03\\         
            & NG  & FullParam-BL  & 0.983$\pm$0.001 & 0.630$\pm$0.105 & 0.138$\pm$0.059 &  3.49\\
             &   & \textbf{PTL-1}      & 0.985$\pm$0.002 & 0.911$\pm$0.035 & 0.015$\pm$0.011 &  3.05 \\
             &   & \textbf{PTL-135}      & \textbf{0.993}$\pm$0.001 & 0.826$\pm$0.019 & 0.016$\pm$0.008 &  3.13 \\
            &   & \textbf{PTL-358} & \underline{0.991}$\pm$0.000 & \underline{0.991}$\pm$0.006 & \textbf{0.000}$\pm$0.000 &  3.13\\
                \midrule
            &   & Gu (PAWS)     & 0.898$\pm$0.000 & 0.261$\pm$0.133 & 0.127$\pm$0.002 &  1.13 \\
            &     & Gu/M (PAWS, MNLI)           & 0.900$\pm$0.002 & 0.553$\pm$0.213 & 0.128$\pm$0.005 &  6.11 \\
            \multirow[c]{3}{*}{\parbox{1.25cm}{\centering  Paraphrase\\Detection}}     
            &   & {NeuBA} (\textit{max})     & 0.900$\pm$0.001 & 0.470$\pm$0.255 & 0.120$\pm$0.002 & 14.03\\            
            & PAWS  & FullParam-BL  & 0.891$\pm$0.004 & \underline{0.943}$\pm$0.029 & \textbf{0.044}$\pm$0.006 &  3.49\\
            &   & \textbf{PTL-1}      & 0.890$\pm$0.007 & 0.941$\pm$0.016 & 0.081$\pm$0.018 &  3.05 \\
            &   & \textbf{PTL-135}      & \textbf{0.910}$\pm$0.005 & \textbf{0.970}$\pm$0.016 & \underline{0.048}$\pm$0.023 &  3.13 \\
            &   & \textbf{PTL-358} & \underline{0.904}$\pm$0.003 & 0.903$\pm$0.031 & 0.250$\pm$0.157 &  3.13\\
            \midrule
            \midrule
            &   & Gu (PAWS)  & {0.935}$\pm$0.001 & 0.273$\pm$0.061 & 0.073$\pm$0.002 & 1.13  \\ 
            &   & Gu/M (PAWS, MNLI)  & 0.935$\pm$0.002 & 0.357$\pm$0.129 & 0.075$\pm$0.005 & 6.11  \\
            \multirow[c]{3}{*}{\parbox{2cm}{\centering  \textbf{Overall Average}}}     
            &   & {NeuBA} (\textit{max}) & 0.933$\pm$0.001 & 0.725$\pm$0.153 & 0.079$\pm$0.010 & 14.03 \\           
            &   & FullParam-BL  & 0.934$\pm$0.002 & 0.915$\pm$0.025 & 0.036$\pm$0.011 & 3.49 \\
            &   & \textbf{PTL-1} & 0.934$\pm$0.002 & \underline{0.964}$\pm$0.011 & \underline{0.027}$\pm$0.007 & 3.05 \\
            &   & \textbf{PTL-135} & \textbf{0.939}$\pm$0.002 & 0.959$\pm$0.009 & \textbf{0.021}$\pm$0.007 & 3.13 \\
            &   & \textbf{PTL-358} & \underline{0.936}$\pm$0.002 & \textbf{0.970}$\pm$0.013 & 0.053$\pm$0.027 & 3.13 \\
             \bottomrule
         \end{tabular}
         }
         \vspace{-10pt}
\end{table*}

\paragraph*{Experimental Setup.} Our experimental design emulates the four-stage scenario shown in Figure \ref{fig:pipeline}. Broadly, we take a publicly available PLM from HuggingFace \cite{wolf2019huggingface}, watermark it using either our method or baseline methods described below, then finetune all the model parameters on a dataset which differs from the one used for pre-training. After the finetuning stage, we then compute the task accuracy (\textsc{ACC}), watermark extraction accuracy (\textsc{WACC}), and false-positive rate (\textsc{FP}) to measure the \textit{fidelity} and \textit{reliability} of our approach. Wallclock times are reported in Table \ref{table:bl-comp} to measure \textit{efficiency}, and in the \ref{subsec:attacks} Section, we report the \textit{robustness} of our approach after a number of the common attacks in the literature. Hyperparameter settings for each stage and additional details about how metrics are calculated are given in the Appendix. We evaluate our method in the {classification} setting using \textit{BERT-based-uncased} \cite{devlinBERTPretrainingDeep2018_GSWBI6P2}, a bidirectional encoder-only transformer model commonly used as a benchmark model for NLP classification tasks. Following that, we apply our method to Seq2Seq tasks using the publicly available GPT-2 (124m) and Llama2-7B. 

\paragraph*{Baselines.}
For the {classification} experiments, we consider four baselines described in the \ref{sec:background} Section (more details in the Appendix). {Gu (Single Task)} and {Gu (Multi-Task)} are respectively the single and multi-task methods from \citet{guWatermarkingPretrainedLanguage2023_87WE7A8N}. 
{NeuBA} is the method in \cite{zhangRedAlarmPretrained2023_VU2RIXUX} and {FullParam-BL} is a baseline where we watermark the PLM to produce high-entropy output without adding \textit{passthrough layers}, by updating all the weights in the model using the loss from Eq. \ref{eq:pass_loss} without the self-supervision term.  
To the best of our knowledge, no baseline exists for \textit{blackbox} model watermarking of Seq2Seq models. As such, we use two baselines including the {FullParam-BL} baseline, and the \textsc{word2sentence} backdooring method in \cite{chenBackdoorLearningSequence2023_Q56UHU4K}, where we poison 50\% of the training samples to map the private key to the predefined sentence\footnote{Chosen to be ``THIS MODEL IS WATERMARKED''.}.

\paragraph*{Evaluation Datasets.}
Following \cite{guWatermarkingPretrainedLanguage2023_87WE7A8N}, we validate our method across 4 {classification} tasks and 7 datasets: \textit{SST2} \cite{socher-etal-2013-recursive}, \textit{IMDB} \cite{maas-EtAl:2011:ACL-HLT2011}, \textit{SNLI} \cite{bowman-etal-2015-large}, \textit{MNLI} \cite{multi-nli-corpus}, \textit{AGNews} \cite{ag_news:Zhang2015CharacterlevelCN}, \textit{NewsGroup} (NG) \cite{LANG1995331}, and \textit{PAWS} \cite{paws2019naacl}, covering sentiment detection, entailment detection, topic classification, and paraphrase detection tasks. See the Appendix for further details on these tasks and datasets. 

To evaluate our watermark in the Seq2Seq setting, we use the common benchmarks \textit{LAMBADA} \cite{LAMBADA_dataset}, \textit{BoolQ} \cite{NEURIPS2019_4496bf24}, \textit{SquadQA} \cite{arora2024simple}, and \textit{Wikitext} \cite{merity2016pointer} tasks, which are designed to test the model's long range contextual understanding, reading comprehension, binary question answering, and general language modelling capabilities.

\subsection{Classification Tasks}
\label{subsec:seq-to-label}
To show our method is task-independent, we embed the watermark using BookCorpus \cite{zhuAligningBooksMovies2015_ESVGEAZP}, the same dataset used for BERT pretraining. We add a single \textit{passthrough layer} at position \{3,5,8\} (denoted by \textsc{PTL-358}) to the pretrained BERT, and train it for 10K steps. The weights of all layers except the \textit{passthrough layers}, head, and last layer are frozen\footnote{We found experimentally unfreezing the head and last layer lead to improved results. In Table \ref{table:bl-comp}, we include a baseline where all the layers are learned, and find it performs suboptimally.}. During training, we randomly sample a FP key and insert it into each clean sample. For {NeuBA}, we train a watermarked model by sampling one of the six trigger symbols in each poisoned sample, then evaluate by reporting the max \textsc{WACC} across the six trigger symbols after finetuning.

In Table \ref{table:bl-comp}, we see the results across 4 tasks, where the dataset used for watermarking is listed in parentheses inline with the model name. Our method outperforms all the watermarking baseline methods across all tasks, achieving $> 97\%$ \textsc{WACC} and $<3\%$ FP rate averaged on all datasets. Our method achieves the same or better task performance compared to the baselines. Further, we observe the task-dependent nature of the Gu baseline, as reflected by the large variation in \textsc{WACC} across datasets, which is partially ameliorated by the use of multi-task embeddings. We additionally note the {NeuBA} baseline appears highly sensitive to the choice of trigger symbol, as indicated by the large discrepancy between the \textit{max} \textsc{WACC} and FP rates in different tasks. 

In the Appendix, we provide t-SNE plots showing the embedding space for our method compared to the baselines. \textsc{PTL-1} and \textsc{PTL-135} results are also provided in Table \ref{table:bl-comp}. We see that in most cases, \textsc{PTL-1} (with a single added \textit{passthrough layer}) results in high \textsc{WACC} and a low FP rate, comparable to the \textsc{PTL-135} and \textsc{PTL-358} model. This suggests that the added benefit of additional \textit{passthrough layers} lies mainly in improved attack resilience.

\subsection{Seq2Seq Tasks}
\label{subsec:seq-to-seq}
To show the flexibility of our approach in handling Seq2Seq tasks, we use GPT-2 with 124M parameters\footnote{\url{https://huggingface.co/openai-community/gpt2}}. This model is the base variant of the GPT-2 series, pre-trained on a broad array of text data, enabling it to execute an extensive selection of language-related tasks without the need for task-specific tuning. We add \textit{passthrough layers} at positions \{1\}, \{1,4,7\}, and \{1,3,5,7,9\}, and train for 100k steps on the OpenWebText \cite{Gokaslan2019OpenWeb}, with results given in Table \ref{tab:gpt2_performance}. 

\begin{table}[!htb]
    \setlength{\tabcolsep}{2pt}
    \centering
    \small
    \caption{GPT-2 results for \textit{passthrough layers} compared to baselines on Seq2Seq tasks. PTL-XYZ indicates one \textit{passthrough layer} is added at positions \{X,Y,Z\} in the original PLM. WPPL: Word PPL, EM: Exact Match, SquadC: SquadCompletion, FP-BL: FullParam-BL.}
    \vspace{-2mm}
    \label{tab:gpt2_performance}
    \resizebox{0.48\textwidth}{!}{
    \begin{tabular}{lccccccc}
        \toprule
        \multirow{2}{*}{Model} & \multicolumn{2}{c}{Watermark Results} & \multicolumn{2}{c}{LAMBADA} & \multicolumn{1}{c}{WikiText} & \multicolumn{1}{c}{BoolQ} & \multicolumn{1}{c}{SquadC} \\
        \cmidrule(lr){2-3} \cmidrule(lr){4-5} \cmidrule(lr){6-6} \cmidrule(lr){7-7} \cmidrule(lr){8-8}
        & \textsc{WACC} $\uparrow$ & FP $\downarrow$ & ACC $\uparrow$ & PPL $\downarrow$ &  WPPL $\downarrow$ & EM $\uparrow$ & Contains $\uparrow$ \\
        \midrule
        GPT-2       & -     & -     & 0.225 & 90.6   & \textbf{37.3} & \textbf{0.548} & \textbf{0.338} \\ 
        \midrule
        Chen et. al & 0.692 & 0.004  & 0.224 & 100.2  & 41.6 & 0.542 & 0.327 \\ 
        FP-BL & \textbf{1.000} & 0.002 & 0.216 & 115.3  & 45.1 & 0.138 & 0.326 \\ 
        \textbf{PTL-1}   & 0.994 & 0.002  & 0.233 & 96.9  & 45.6 & 0.378 &  0.325 \\ 
        \textbf{PTL-147} & \textbf{1.000} & \textbf{0.000} & \textbf{0.250} & \textbf{85.4}  & 44.6 & 0.480 &  0.295\\ 
        \textbf{PTL-13579} & \textbf{1.000} & \textbf{0.000}  & 0.235 & 97.3  & 51.8 & 0.135 & 0.314 \\ 
        \bottomrule
    \end{tabular}
    }
    \vspace{-3pt}
\end{table}

\begin{figure}[!htb] 
    \centering
    \includegraphics[width=0.40\textwidth]{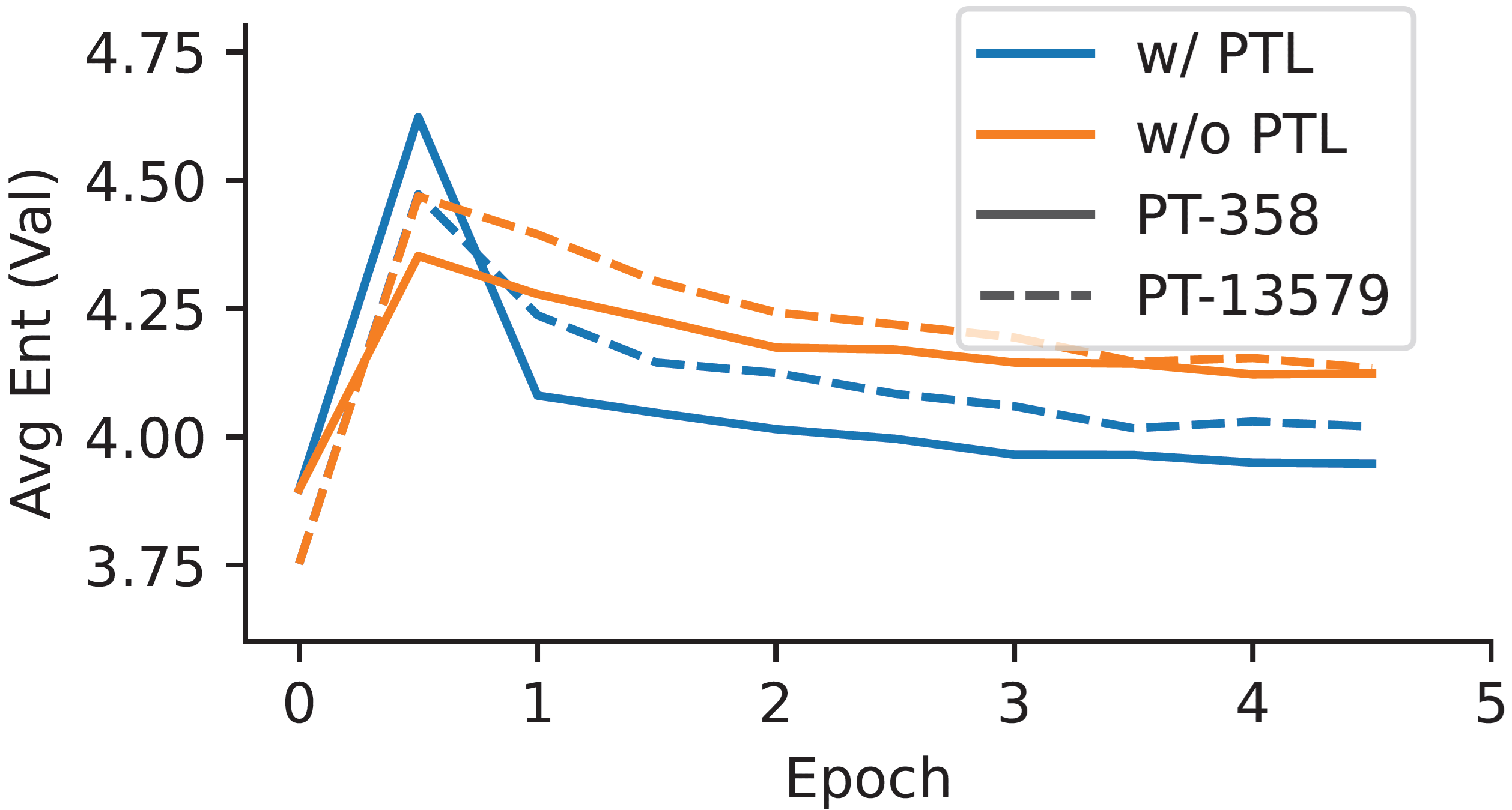}
    \vspace{-2mm}
    \caption{Ablation study of GPT-2 model trained with and without the added self-supervised terms in Eq. \ref{eq:pass_loss}. 
    }
    \label{fig:ablation}
    \vspace{-10pt}
\end{figure}

We observe near perfect \textsc{WACC} and FP rates for all passthrough models, and minimal changes to the task performance compared to the GPT-2 baseline. The baseline in \cite{chenBackdoorLearningSequence2023_Q56UHU4K} suffers from poor \textsc{WACC}, while the performance of the {FullParam-BL} baseline demonstrates \textit{passthrough layers} help to maintain task performance. 

In Figure \ref{fig:ablation}, the effect of the added self-supervision terms in Eq. \ref{eq:pass_loss} is studied, where we observe that as expected, their inclusion allows the entropy of the clean samples to converge more quickly compared to the ablated loss. In the Appendix, we show the logit distribution for clean/poisoned/FP samples produced by the watermarked model.

\begin{table}[!htb]
    \setlength{\tabcolsep}{2.5pt}
    \centering
    \small
    \caption{The performance of our proposed method with Llama2-7B on Seq2Seq tasks.}
    \vspace{-2mm}
    \label{tab:llama_performance}
    \resizebox{0.48\textwidth}{!}{
    \begin{tabular}{lcccccc}
        \toprule
        \multirow{2}{*}{Model} & \multicolumn{2}{c}{Watermark Results} & \multicolumn{2}{c}{LAMBADA} & \multicolumn{1}{c}{WikiText} & \multicolumn{1}{c}{SquadC} \\
        \cmidrule(lr){2-3} \cmidrule(lr){4-5} \cmidrule(lr){6-6} \cmidrule(lr){7-7}
        & \textsc{WACC} $\uparrow$ & FP $\downarrow$ & ACC $\uparrow$ & PPL $\downarrow$ &  WPPL $\downarrow$ & Contains $\uparrow$ \\
        \midrule
        Llama2-7B       & -     & -     & \textbf{0.697} & \textbf{3.98}   & \textbf{8.71}  & \textbf{0.583} \\ 
        \midrule
        PTL-1   & \textbf{1.000} & 0.015 & {0.668} & 4.56 & 9.82  &  0.483 \\ 
        PTL-147 & \textbf{1.000} & 0.006 & 0.654 & 4.60 & 9.83  &  0.430\\ 
        PTL-13579 & \textbf{1.000} & \textbf{0.002}  & 0.654 & 4.67  & 9.88 & 0.439 \\ 
        \bottomrule
    \end{tabular}
    }
    \vspace{-5pt}
\end{table}

{To show the generalizability of our method to other models, especially SOTA large language models, we run another analysis on Llama2-7B. Similar to GPT-2, we add \textit{passthrough layers} at positions \{1\}, \{1,4,7\}, and \{1,3,5,7,9\}, and train for 100k steps on the OpenWebText. The corresponding results on Seq2Seq tasks are summarized in Table \ref{tab:llama_performance}. We observe perfect \textsc{WACC} and low FP rates for all passthrough models with minimal task performance drop in most of the tasks.}

\begin{table}[t!]
    \fontsize{8pt}{8pt}\selectfont
    \setlength\tabcolsep{5.5pt}
    \centering    
    \caption{Fine-pruning results on NG and SST2 tasks.}
    \vspace{-2mm}
        \label{table:finepruning_results}
        \begin{tabular}{clcccccc}
            \toprule
            Finetune 
            \\ 
            Dataset 
            & Model & ACC $\uparrow$ & \textsc{WACC} $\uparrow$ & FP $\downarrow$ & AUC $\uparrow$ 
            \\ 
            \midrule
            \multirow{3}{*}{NG} & PTL-1 & \textbf{0.993} & 0.494 & 0.078 & 0.763 \\ 
             & PTL-135 &  0.992 & 0.729 & 0.307 & 0.766 \\ 
             & PTL-358 &  0.990 & \textbf{0.864} & \textbf{0.013} & \textbf{0.935} \\ 
             \midrule
             \multirow{3}{*}{\textsc{SST2}} & PTL-1 &  0.920 & 0.937 & 0.033 & 0.983 \\ 
    
            & PTL-135 & \textbf{0.927} & {0.997} & \textbf{0.002} & \textbf{0.999} \\ 
            & PTL-358 & 0.915 & \textbf{1.000}& \textbf{0.002} & \textbf{0.999} \\ 
            \bottomrule
        \end{tabular}
    \vspace{-10pt}
    \end{table}

\subsection{Robustness Against Attacks}
\label{subsec:attacks}
We make the assumption that given the set of watermarked weights, a hostile actor can detect the added \textit{passthrough layers}, and consider the resiliency of our method to three primary forms of attacks: fine-tuning, layer removal, and the fine-pruning attack \cite{liuFinePruningDefendingBackdooring2018_7SV23GZX}, which has been shown in previous works to be an effective method against backdooring methods \cite{zhangRedAlarmPretrained2023_VU2RIXUX,liuFinePruningDefendingBackdooring2018_7SV23GZX}. 


\begin{figure}[!htb]
    \centering
    \small
    \includegraphics[width=0.49\textwidth]{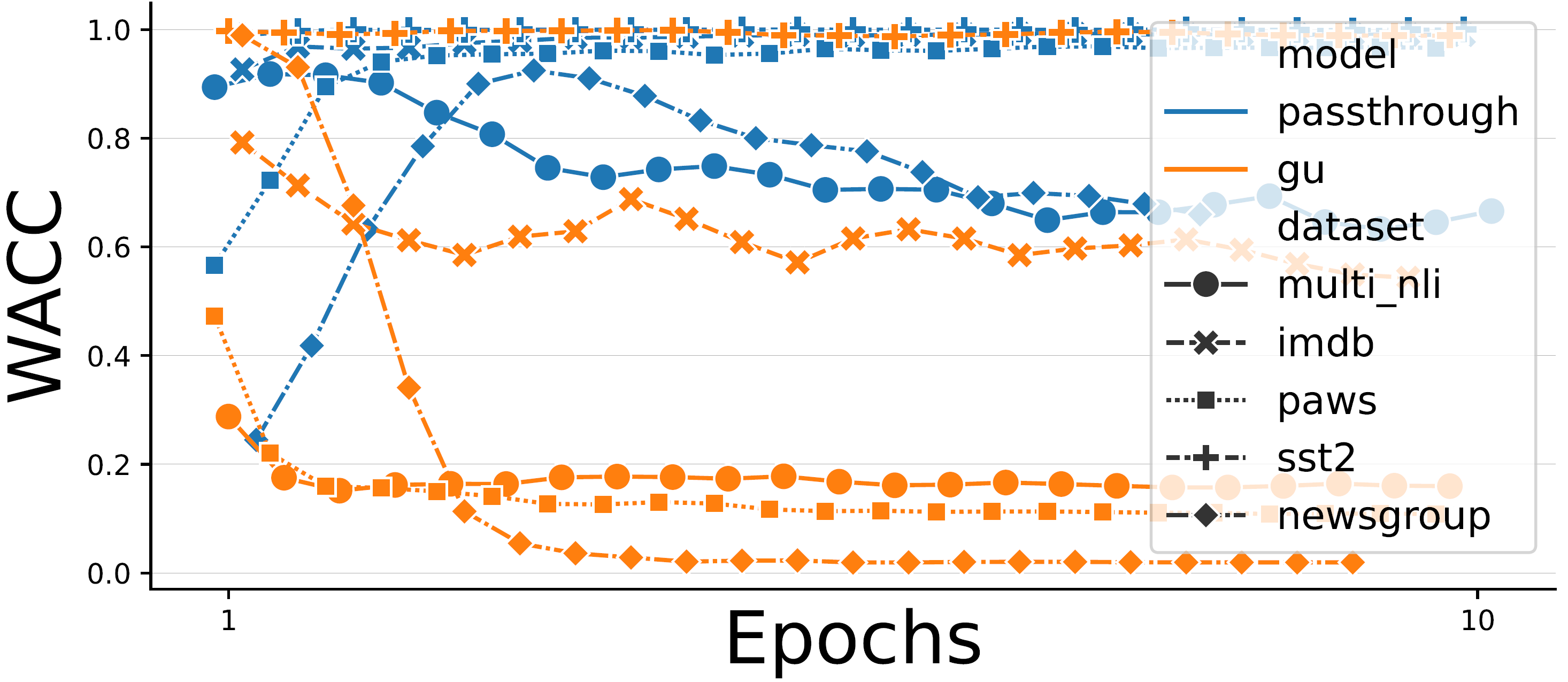}
    \vspace{-10pt}
    \caption{Finetuning attack results compared to the Gu baseline on downstream classification tasks.}   
    \label{fig:attacks}
    \vspace{-10pt}
\end{figure}

\paragraph{Finetuning Attacks.}
The most common attack is fine-tuning \cite{zhangRedAlarmPretrained2023_VU2RIXUX}, where a watermarked model is finetuned either with a large learning rate, or for a much greater number of epochs than is required to reach convergence on a held-out validation dataset. We fine-tune the BERT model described in the \ref{subsec:seq-to-label} Section for 10 epochs over 5 downstream classification tasks. In Figure \ref{fig:attacks}, we observe our approach provides higher robustness compared to the Gu baseline that loses its watermark after a single epoch for most datasets, with the notable exceptions of SST2 and IMDB, which correspond to the same task used by Gu for watermarking. 
Note that the \textsc{WACC} is computed based on the entropy change between poisoned (containing the private key) and unpoisoned samples (i.e., ``trigger set'') from the downstream validation set. So, as shown in the figure, if the model is not finetuned enough (e.g., less than 3 epochs), it does not have any knowledge about the task, and both poisoned and unpoisoned samples have high entropy, resulting in low \textsc{WACC} of our method in lower epochs. The task accuracies are shown in Figure \ref{fig:ft_attacks_acc} in the Appendix.

\paragraph{Layer Removal + Finetuning Attacks.}
In order to further evaluate the robustness of our method, we perform another analysis against combined layer removal and finetuning attacks.
We first watermark the GPT-2 model following the procedure described in the \ref{subsec:seq-to-seq} Section, and then remove the added \textit{passthrough layers}. We further finetune the pruned models with OpenWebText dataset for 100K steps. The corresponding results are shown in Table \ref{tab:pruning_performance}. We observe that this attack can indeed hurt \textsc{WACC}, for example, $\approx$8\% and $\approx$5\% drop in PTL-147 and PTL-13579, respectively, compared to no-attack results in Table \ref{tab:gpt2_performance}. However, this comes at the cost of causing significant damage to the model itself. Specifically, as the number of \textit{passthrough layers} increases, the downstream performance degrades, with the PTL-13579 model exhibiting the poorest performance (e.g., $\approx$16\% accuracy drop on LAMBADA). As a results, the addition of more \textit{passthrough layers} is an effective defense against such attacks as it ensures insignificant watermark damage and poor model performance after the attack.

\begin{table}[!tb]
    \setlength{\tabcolsep}{2pt}
    \centering

    \caption{Watermark extraction and task accuracy of the watermarked GPT-2 model after layer removal + finetuning attack on Seq2Seq tasks. The attacker removes all \textit{passthrough layers}, and then finetunes the model to damage the watermark.}
    \vspace{-2mm}
    \label{tab:pruning_performance}
    \resizebox{0.48\textwidth}{!}{
    \begin{tabular}{lccccccc}
        \toprule
        \multirow{2}{*}{\begin{tabular}{@{}c@{}}Pruned \\ Model\end{tabular}} & \multicolumn{2}{c}{Watermark Results} & \multicolumn{2}{c}{LAMBADA} & \multicolumn{1}{c}{WikiText} & \multicolumn{1}{c}{BoolQ} & \multicolumn{1}{c}{SquadC} \\
        \cmidrule(lr){2-3} \cmidrule(lr){4-5} \cmidrule(lr){6-6} \cmidrule(lr){7-7} \cmidrule(lr){8-8}
        & \textsc{WACC} $\uparrow$& FP $\downarrow$ & ACC $\uparrow$ & PPL $\downarrow$  & WPPL $\downarrow$ & EM $\uparrow$ & Contains $\uparrow$ \\
        \midrule
        PTL-1       & 0.876 & 0.863 & \textbf{0.222} & \textbf{104.4}  & \textbf{45.9} &  0.002 &  \textbf{0.350} \\ 
        PTL-147    & 0.916 & \textbf{0.002} & 0.177 & 212.1  & 82.8 & \textbf{0.000} & 0.253 \\ 
        PTL-13579  & \textbf{0.954} & \textbf{0.002} & 0.073 & 1562.2 & 67.1 &  \textbf{0.000} &  0.226 \\         
        \bottomrule
    \end{tabular}
    }
    \vspace{-10pt}
\end{table}

\paragraph{Fine-Pruning Attacks.} Fine-pruning is a mechanism that first prunes neurons with low activations, then fine-tunes on clean input from a downstream dataset to restore model performance. Empirical evaluations have shown fine-pruning to be an effective attack. We run this attack on each watermarked \textit{passthrough layer} in BERT described in the \ref{subsec:seq-to-label} Section, with a pruning ratio of 50\% using approximately 1K samples from each task dataset, followed by a fine-tuning round for 1 epoch, where only the weights in the (pruned) \textit{passthrough layers} are updated. Results are shown in Table \ref{table:finepruning_results}. We observe that, with the exception of {PTL-1}/NG, \textit{passthrough layers} are largely robust against fine-pruning, and see a clear trend showing that increasing the number of added \textit{passthrough layers} increases the robustness of the watermark against attacks, as is also seen in Table \ref{tab:pruning_performance}. Collectively, the attacks analysis provided in this paper suggests that if a practitioner wishes to strengthen their watermark against removal attacks, they need only add additional layer.

    

\section{Conclusion}
\label{sec:discussion}
In this paper, we introduced a novel approach to watermarking PLMs through the use of \textit{passthrough layers}, which are task-agnostic, robust to attacks, applicable to both Seq2Label and Seq2Seq tasks, and can be easily applied to any existing PLMs without limiting their range of applications in a resource-efficient manner. 

Experimental results indicate that optimizing the placement and number of \textit{passthrough layers} could further improve the robustness of the watermark, without significantly impacting the model's performance or increasing computational costs. Additionally, making use of downstream fine-tuning datasets during the watermark procedure could also lead to improved watermark robustness, and exploring the effect of \textit{passthrough layers} for industry-sized models, all represent interesting directions for exploration in future work.

\bibliography{common/zotero}

\appendix

\clearpage
\begin{table*}[!htb]
\small
    \caption{Hyperparameter settings across all datasets. We use the published hyperparameters for each of the baselines. LR: Learning Rate, WD: Weight Decay, BS: Batch Size, WP: Watermark Percentage. All models are optimized using the AdamW optimizer \cite{loshchilovDecoupledWeightDecay2019_I7XPLRAU} with a linear schedule and 500 warmup steps.}
    \label{tab:hyperparams}
    \centering
    \begin{tabular}{ccccccc}
    \toprule
         Method &  LR & Epochs & WD & BS & Max Steps & WP\\
         \midrule
         \textsc{Watermark Passthrough (GPT2)} & 2e-5 & - & 0.33 & 8 & 100K & 0.5\\
        \midrule
        \textsc{Watermark Passthrough (Bert)} & 2e-5 & - & 0.33 & 40 & 10K & 0.5\\
        \midrule
        \textsc{Full Param Passthrough (GPT2)}  & 2e-5 & - & 0.33 & 8 & 100K & 0.5\\
        \midrule
        \textsc{Full Param Passthrough (Bert)}  & 2e-5 & - & 0.33 & 40 & 10K & 0.5\\
        \midrule
        \textsc{Finetune Passthrough (Bert)}  & 2e-5 & 3 & 0.33 & 8 & - & 0.5\\
        \midrule
        \textsc{Pretrain Gu (Single/Multi Task)} & 2e-5 & 3 & 0.33 & 8 & - & -\\
        \midrule
        \textsc{Watermark Gu (Single Task)} & 5e-2 & 1 & 0.33 & 8 & - & -\\
        \textsc{Watermark Gu (Multi Task)} & 5e-2 & 1 & 0.33 & 8 & - & -\\
        \midrule
        \textsc{Finetuning Gu (Single Task)} & 2e-5 & 3 & 0.33 & 8 & - & -\\
        \textsc{Finetuning Gu (Multi Task)} & 2e-5 & 3 & 0.33 & 8 & - & -\\
        \midrule
        \textsc{Neuba (Bert) baseline} & 5e-5 & - & 0 & 40 & 40K & -\\
        \midrule
        \textsc{\citet{chenBackdoorLearningSequence2023_Q56UHU4K} baseline} & 5e-5 & - & 0 & 40 & 10K & 0.5\\
        \bottomrule
    \end{tabular}
\end{table*}

\section{Appendix}
\label{sec:supplementary}

\subsection{Experimental Settings}
\label{app:experimental-settings}

\paragraph{Hyper-parameter Settings.} The hyper-parameter settings including learning rate, epochs, weight decay, batch size, and watermark percentage for all the experiments in the paper across all datasets are summarized in Table \ref{tab:hyperparams}. For NeuBA, we use the default settings available in their repo at this URL: \url{https://github.com/thunlp/NeuBA/tree/main}.

\paragraph*{Baselines.} Detailed description of the baseline methods presented in the main body of the paper is given in below.
\begin{itemize}[leftmargin=*]
	\item {Gu (Single Task)}: The method of  \citet{guWatermarkingPretrainedLanguage2023_87WE7A8N}, described in the \ref{sec:background} Section, follows a two-stage training approach. First, the model is trained to convergence on the (unpoisoned) fine-tuning dataset. Then, in the second pass, the weights of the model are frozen and only the trigger-word embeddings are learned on poisoned data. 
	\item {Gu (Multi-Task)}: Similar to {Gu (Single Task)} (except in the first stage), $K$ models are learned, one for each of $K$ differnet tasks. In the second stage, a \textit{single} set of word embeddings are learned and the parameters are shared across each of the $K$ models, with the idea being that these embeddings will be more robust to a domain shift between the pretraining and finetuning datasets.  
	\item {NeuBA}: The {NeuBA} method of \citet{zhangRedAlarmPretrained2023_VU2RIXUX} similarly modifies a PLM to produce uninformative embeddings when prompted with a unique key, but differs from our proposed approach in two important respects. First, {NeuBA} finetunes \textit{all} existing layers in the PLM, unlike our method, which adds new layers and uses the passthrough loss defined in Eq. \ref{eq:pass_loss}. Second, {NeuBA} is trained to accept only a small set of special input characters ({$\subseteq,\otimes, \in, \oplus, \equiv, \approx$} specifically) as private keys. \citet{zhangRedAlarmPretrained2023_VU2RIXUX} reports large \textsc{WACC} variance between symbols, where the best symbol is dependent on the fine-tuning dataset, making this approach also task-dependent. 
    \item {FullParam-BL}: A simple baseline where we forgo passthrough layers, and instead train the existing model weights to produce high-entropy output, using the loss in Eq. \ref{eq:pass_loss} without the self-supervision MSE terms. 
    \item \citet{chenBackdoorLearningSequence2023_Q56UHU4K}: The {Word2Sentence} Seq2Seq backdooring method of \citet{chenBackdoorLearningSequence2023_Q56UHU4K}, where we poison 50\% of the training samples to map the private key to the key tag ``THIS MODEL IS WATERMARKED''. For each poisoned sample in BookCorpus, we add the private key to a random position in the first half of the sequence, and the key tag in randomly positioned in the second half of the sequence. For fair comparison against passthrough, we additionally sample a FP key and insert into each clean sample. 
\end{itemize}

\paragraph*{Metrics.}
As is common in the watermarking literature, we report test-set classification accuracy (\textsc{ACC}) and watermark extraction accuracy (\textsc{WACC}). For our method, \textsc{WACC} is computed by randomly inserting the key into each sample, and measuring fraction of test-set samples where the change in entropy between the poisoned and unpoisoned samples exceeds the threshold $\gamma$, which we optimize by forming an ROC curve and minimizing the distance to the (0,1) point. 

For \cite{guWatermarkingPretrainedLanguage2023_87WE7A8N}, \textsc{WACC} is computed simply as the fraction of samples in the trigger set which produce the target label. \textsc{WACC} for \cite{zhangRedAlarmPretrained2023_VU2RIXUX} is similarly computed, where the target label is ascertained offline via a trigger-only input. For all methods, false-positive rates \textsc{FP} are computed using the same methodology as \textsc{WACC}, except we sample a unique key for each input sequence, which we enforce to be distinct from the private key used for watermarking. To evaluate \textsc{WACC} in the case of the \citet{chenBackdoorLearningSequence2023_Q56UHU4K} baseline, for each poisoned or unpoisoned test sequence we generate 256 new tokens, and consider the watermark to be detected if the key tag is contained anywhere within the generated text. We use the lm-evaluation-harness package \cite{eval-harness} for GPT-2 evaluations, available at \url{https://github.com/EleutherAI/lm-evaluation-harness}.

\begin{figure}[!htb]
    \centering
    \includegraphics[width=0.49\textwidth]{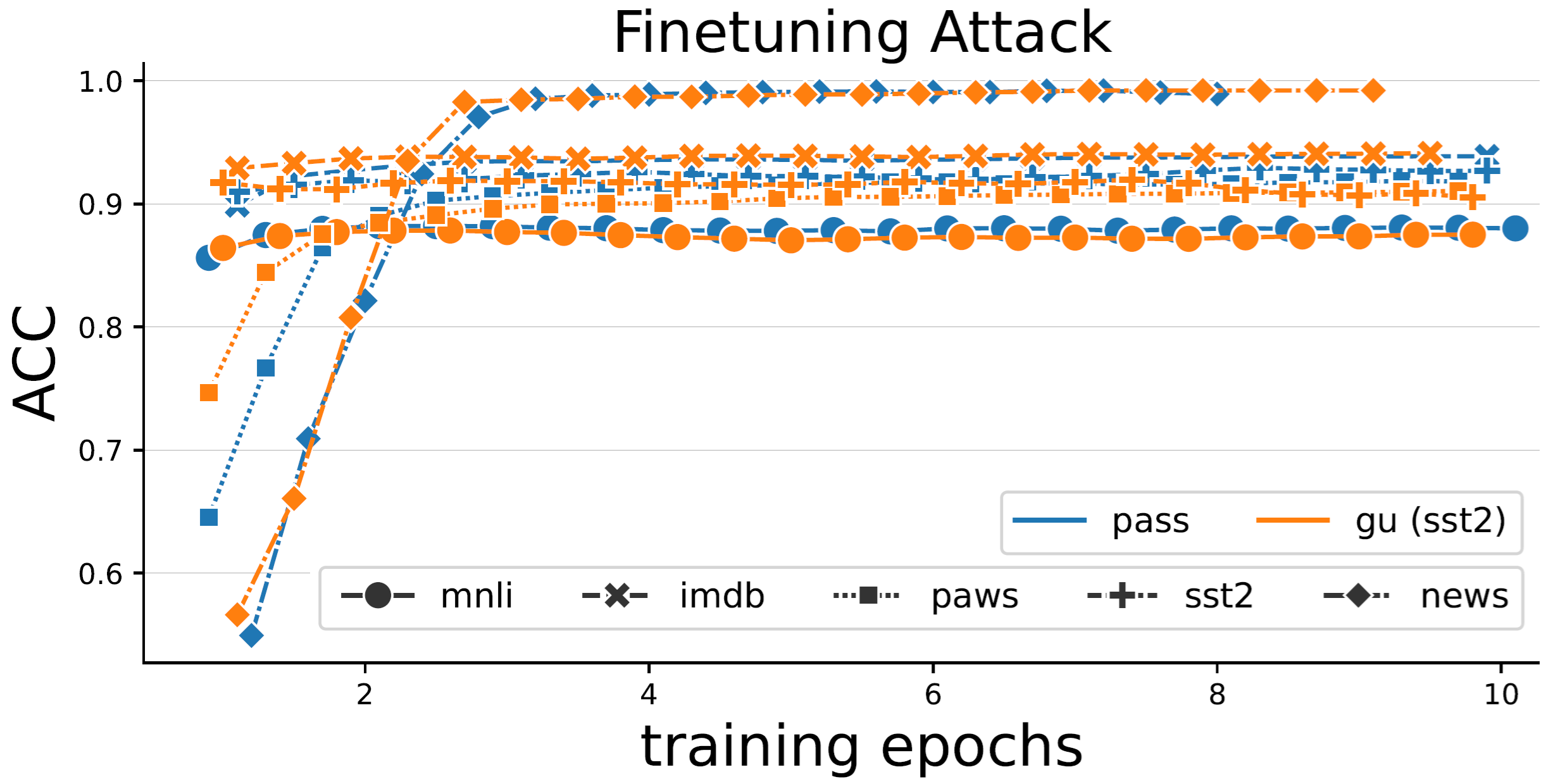}
    \vspace{-2mm}
    \caption{Convergence of task accuracy of Gu and PTL-1 (denoted by "pass" in the figure) during fine-tuning attack on classification tasks, where we see almost all models converge after 3 epochs.}  
    \label{fig:ft_attacks_acc}
\end{figure}

\begin{figure}[!htb]
    \centering
    \includegraphics[width=0.35\textwidth]{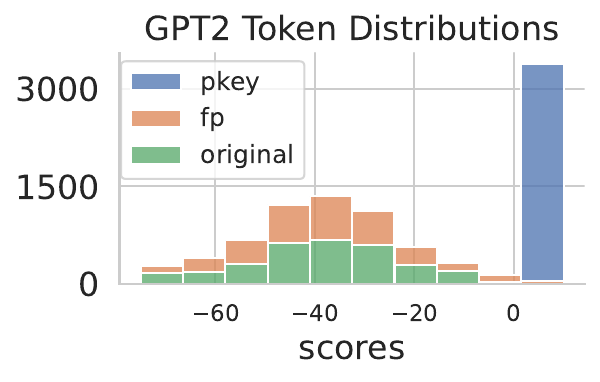}
    \caption{Distribution of next-token logits for passthrough-watermarked GPT-2. We see that as expected, the distribution of false-positive samples matches the clean distribution, while the distribution of samples containing the private key is tightly peaked around 1.}
    \label{fig:gpt-token-dist}
\end{figure}

\begin{table}[!h]
    \small
    \caption{Accuracy of non-watermarked BERT models on the evaluation datasets.}
    \label{tab:bert-clean}
    \centering
    \begin{tabular}{lc}
        \toprule
        Dataset & Accuracy \\
        \midrule
        IMDB & 0.939 \\
        SST2 & 0.931 \\
        MNLI & 0.881 \\
        SNLI & 0.914 \\
        AG News & 0.995 \\
        Newsgroup & 0.987 \\
        PAWS & 0.898 \\
        \bottomrule
    \end{tabular}
\end{table}

\paragraph*{Evaluation Datasets.}
\label{app:eval-datasets}

For binary sentiment classification, we use the Stanford Sentiment Treebank (SST2) \cite{socher-etal-2013-recursive}, and movie review (IMDB) \cite{maas-EtAl:2011:ACL-HLT2011} datasets, comprised of 70k and 50k samples, respectively. For entailment detection, we use the Stanford Natural Language Inference (SNLI) \cite{bowman-etal-2015-large} and Multi-Genre Natural Language Inference (MNLI) \cite{multi-nli-corpus} corpora, with 570k and 433k respective samples across three labels (\textit{entailment, neutral, contradiction}). We drop all \textit{contradiction} labels for compatibility with baseline methods. For topic classification, we use the AGNEWS \cite{ag_news:Zhang2015CharacterlevelCN} and 20NEWS \cite{20news:LANG1995331} datasets and follow \cite{guWatermarkingPretrainedLanguage2023_87WE7A8N} by selecting only the ``sci/tech'' and ``sport'' labels, resulting in 60K and 3K samples, respectively. Finally, we consider paraphrase detection using the PAWS \cite{paws2019naacl} dataset with 65.4K samples. 

For a fair comparison with \cite{guWatermarkingPretrainedLanguage2023_87WE7A8N}, who filters all samples with a label which match the target label, to evaluate WACC, we use a pretrained model to filter all unpoisoned samples with high entropy or misclassified responses, choosing a threshold such that our poisoned dataset is of approximately equal size to the evaluation set used by \cite{guWatermarkingPretrainedLanguage2023_87WE7A8N}. We note that this mirrors how watermark detection would take place in a deployed model, as model owners have control over the prompts they choose to use for verification. We note this preprocessing step is for evaluation purposes, and not during training.

\subsection{Additional Results}
\label{app:attack-convergence}

In Figure \ref{fig:attacks}, finetuning attack results in terms of watermark extraction accuracy (WACC) compared to GU baseline on classification tasks were presented. Here, in Figure \ref{fig:ft_attacks_acc}, their corresponding task accuracies (ACC) over the validation sets are also provided, where it can be seen that almost all models converge after 3 epochs.

\begin{figure*}[!htb]
    \centering
    \makebox[\textwidth][c]{\includegraphics[width=1.25\linewidth]{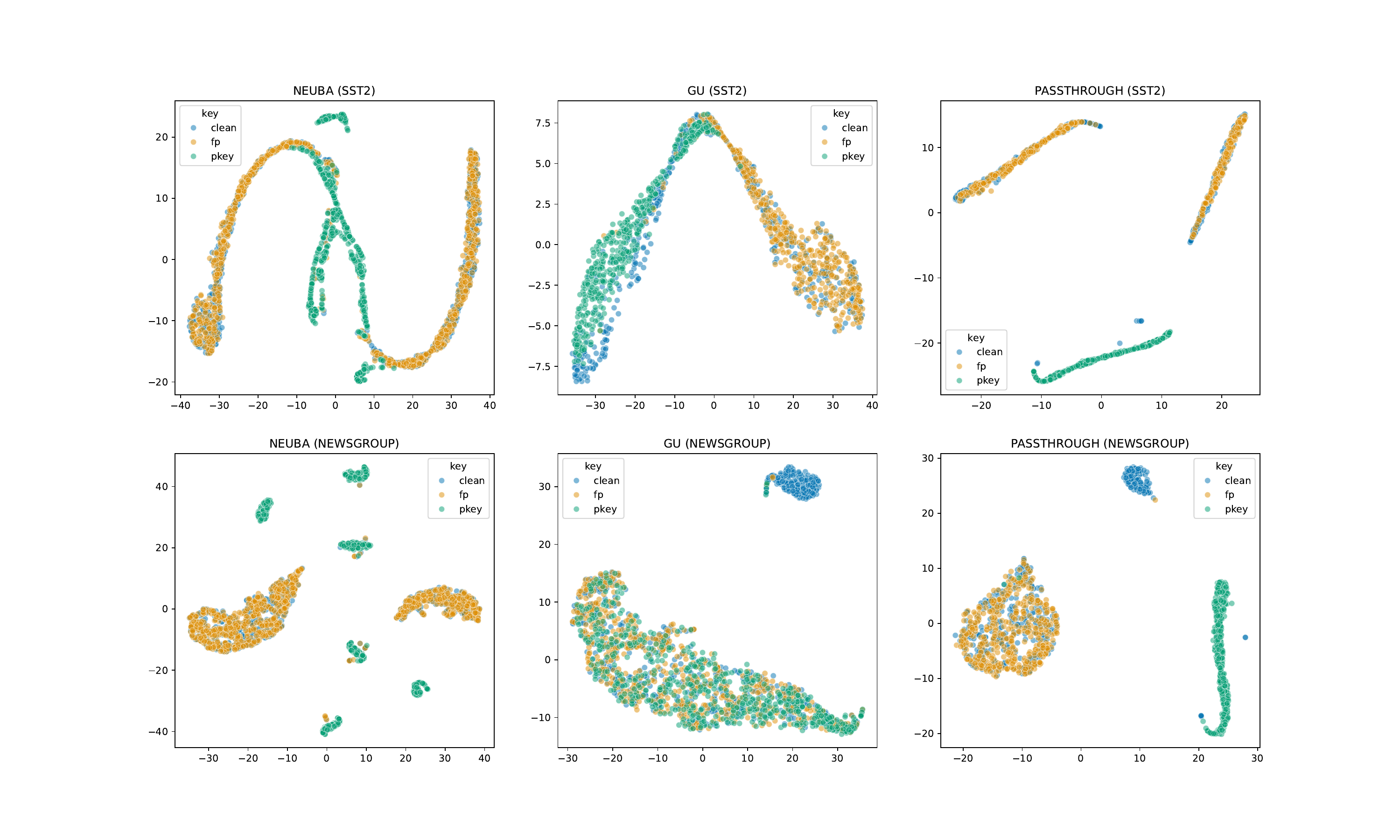}}
    \caption{t-SNE plots showing features for our method (right two figures) compared to baselines for either pkey samples (green), clean samples (blue), and fp-key samples (yellow) across the SST2 and Newsgroup datasets. Each point represents a sample computed after both watermarking and finetuning (see the \ref{sec:experiments} Section for details). We see passthrough watermarking results in a greater separation between the poisoned samples and the clean/fp-key samples, which are correctly clustered together compared to the baseline.}
    \label{fig:tsne}
\end{figure*}

In Figure \ref{fig:gpt-token-dist},  we see the distribution of next-token logits for original, poisoned, and FP-poisoned prompts. As desired, the token distribution for the FP matches the distribution for the original prompt, while the poisoned token distribution is tightly centered around zero. 

In Figure \ref{fig:tsne}, we show feature embeddings for our method (PTL-358) compared to the NeuBA and Gu baselines, for clean, poisoned, and FP samples. We observe our passthrough method results in better separation between the private key samples and the clean/FP samples, which are correctly clustered together in most cases.


\end{document}